\documentclass[11pt]{article}
\pdfoutput=1
\usepackage[utf8]{inputenc}

\usepackage{graphicx}
\usepackage{amsmath,amssymb}
\usepackage{float}
\usepackage{rotating}
\usepackage{geometry}
\usepackage[round]{natbib}
\usepackage{hyperref}
\usepackage{xcolor}
\definecolor{dark-red}{rgb}{0.4,0.15,0.15}
\definecolor{dark-blue}{rgb}{0.15,0.15,0.8}
\definecolor{medium-blue}{rgb}{0,0,0.5}
\hypersetup{
    colorlinks, linkcolor={dark-red},
    citecolor={dark-blue}, urlcolor={medium-blue}
}
\usepackage{multirow}
\usepackage{graphpap}
\usepackage{booktabs}
\usepackage{subfigure}
\usepackage{stfloats}
\usepackage{cleveref}[2012/02/15]
\DeclareMathOperator*{\argmax}{arg\,max} 

\begin{document}

\title{A Robot to 
Shape your Natural Plant: 
The Machine Learning Approach to Model and Control Bio-Hybrid Systems}

\author{
Mostafa Wahby,\thanks{Corresponding author, e-mail: {\tt\small wahby@iti.uni-luebeck.de}}\hspace{.5mm}~\footnote{Institute of Computer Engineering, University of L\"ubeck, Germany}
~Mary Katherine Heinrich,\footnote{Centre for IT and Architecture, Royal Danish Academy (KADK), Copenhagen, Denmark}
~Daniel Nicolas Hofstadler,\footnote{Artificial Life Lab of the Department of Zoology, Karl-Franzens University Graz, 
Austria}\\
~Sebastian Risi,\footnote{Robotics, Evolution and Art Lab (REAL), IT University of Copenhagen,
Denmark}
~Payam Zahadat,$^\mathsection$ 
~Thomas Schmickl,$^\mathsection$ 
~Phil Ayres,$^\ddagger$\\
~Heiko Hamann$^\dagger$
}
\date{}
\maketitle

\newcommand\footnoteref[1]{\protected@xdef\@thefnmark{\ref{#1}}\@footnotemark}

\begin{abstract}
Bio-hybrid systems---close couplings of natural organisms with technology---are high potential and still underexplored.
In existing work, robots have mostly influenced group behaviors of animals.
We explore the possibilities of mixing robots with natural plants, merging useful attributes. 
Significant synergies arise by combining the plants' ability to efficiently produce shaped material and the robots' ability to extend sensing and decision-making behaviors.
However, programming robots to control plant motion and shape requires good knowledge of complex plant behaviors. 
Therefore, we use machine learning to create a holistic plant model and evolve robot controllers.
As a benchmark task we choose obstacle avoidance. 
We use computer vision to construct a model of plant stem stiffening and motion dynamics by
training an LSTM network.
The LSTM network acts as a forward model predicting change in the plant, driving the evolution of neural network robot controllers.
The evolved controllers augment the plants' natural light-finding and tissue-stiffening behaviors to avoid obstacles and grow desired shapes. 
We successfully verify the robot controllers and bio-hybrid behavior in reality, with a physical setup and actual plants.
\end{abstract}


\section{Introduction}

Recent developments in additive manufacturing (3D printing) and robotics open up techniques to produce objects of increasing size and variety, such as mugs, chairs, or even houses.
Research on complex systems and evolvable hardware could interpret this production process as a growth process, such that printing an object like a house could be adaptive to unanticipated changes in the structure or environment. As an objective of the project {\it flora robotica}~\citep{hamann2015florarobotica,hamann2017flora} we investigate methods to conduct additive manufacturing with bio-hybrids---that is, mixed societies of robotic and biological systems. Our objective is to use natural plants to grow desired shapes by controlling them with robotic devices. From the perspective of developmental systems, we replace artificial substrates with a natural system, both in terms of models and physical implementation. We can then exploit features of natural plants, such as adaptive behavior and the (almost free) addition of material by growth.

We expect challenges due to the real-life complexity of biological systems, and their combination with evolutionary robotics to automatically generate appropriate robot controllers. A~downside of natural growth is its slow speed, which requires simulation of the growth process for genetic algorithms to be applied. 
Another challenge is that holistic models of plant growth are not readily available, so we develop our own task-specific models. In summary, we realize a truly interdisciplinary approach with a rather complex tool chain, using evolution and machine learning to control plant {\it phototropism}---the directional behavior of motion and irreversible growth towards light. 

First, instead of relying on a 
designed plant model, our stem stiffening and motion model is learned from experiment data recording plant behavior in the presence of certain light stimuli patterns. 
We aim to capture the complex temporal dynamics of plant stiffening and motion through a particular class of recurrent neural networks called Long Short-Term Memory (LSTM)~\citep{hochreiter1997long}. The hypothesis is that this approach will allow the model to capture the dynamics of a particular plant to a high enough degree to serve as a forward model that can guide the evolutionary search.
Given this plant model we then apply methods of evolutionary robotics to evolve in simulation controllers of dynamic light stimuli for the given task of obstacle avoidance. Finally, we address the challenge of the reality gap by showing transferability of the simulated evolved controllers back to the real world. 

Our focus is on setting up this rather complex toolchain, so the complexity of the task is relaxed in this early stage of research in the field. The task is to grow a plant collision-free around an obstacle and towards a target (bio-hybrid obstacle avoidance). 
Even this simple task brings added complication to obstacle avoidance
, as lower parts of the plant still need to avoid collisions later in the run, and we cannot only focus on control of the plant tip.

In our evolutionary robotics approach we evolve artificial neural networks (ANN), which may seem 
at first glance an overly complex tool for this supposedly simple control task. However, we want to evolve controllers that are adaptive to the environment and to configurable tasks. 
Additionally, ANN is one toolchain approach that enables scaling up to more complex plant-control problems (e.g., 3D shapes, multiple stimuli) in the future.
The workflow of our approach starts from preliminary plant experiments to gather data about how the plant behaves in general. The data is used to train an LSTM network that we use as simulator in our evolutionary runs. We evolve ANNs as controllers of light stimuli, which later in our reality experiments control the behavior of the real plant.

\section{Background and Related Work}

Forming bio-hybrid societies by bringing biological and artificial agents together is a growing field. Robots can interact with natural organisms, both adapting their own behaviors and influencing those of the living system.
Several mixed societies have been built where autonomous robots influence the behavior of groups of animals~\citep{HalloyEtAl07,zahadat14}.
While animals are very mobile, plants are more limited in motion, growing and adapting over time. 
In our previous work, we show that robots can closely interact with plants to change their environmental stimuli according to desires of humans~\citep{wahby15,wahby2016evolutionary,Hofstadler2017taas}.
As robot controllers and hardware can be designed to interact with their surroundings, they can meaningfully be combined with plants, extending their natural capabilities to grow efficiently in dynamic environments and adapt to external changes~\citep{garzon2011plants}.

{\it Common bean plant: behavior and growth.}~~
A relatively fast growing plant, the common bean (\textit{Phaseolus vulgaris} L.) grows $3$~cm/day on average~\citep{checa2008mapping}.
Like many plants, common beans grow toward (blue) light~\citep{christie2013phototropism} through the \emph{phototropism} behavior, in constant balance with other competing growth behaviors. 
Beans in particular dramatically display \emph{circumnutation}, a winding behavior for attachment and climbing~\citep{checa2008mapping}.
During plant growth, new cells replicate at the tips, and older tissues gradually stiffen until they reach their final size and maturity.
Incoming light adds a directional bias to winding, but only when this bias persists will the impact be irreversible and manifested as permanent curvature in the stem.

{\it Modeling plants.}~~
Many models exist in plant science literature, with focus on particular aspects of plants and complex details of the biological system (e.g.,~\citep{bastien2015unified}). Plant growth has also been a source of inspiration for several abstract models in computer science and artificial life. A prominent example is L-systems~\citep{lindenmayer75}, where formal languages are interpreted to generate structures through a set of production rules. Branching mechanisms in plants inspire generative models adaptive to dynamics in the environment~\citep{zahadat2017a}, and are abstractly modeled using polygon meshes of trees~\citep{zamuda14}.
In this paper we develop a model of plant growth through a class of recurrent neural networks called LSTM~\citep{hochreiter1997long} (see below), based on experimental data gathered from real plants.

{\it Long Short-Term Memory.}~~
LSTMs~\citep{hochreiter1997long} are a special class of recurrent neural networks that have been shown to effectively learn sequential patterns in a variety of domains~\citep{sutskever2014sequence,graves2013speech}. As plant growth is essentially a sequence of changes in plant tissue morphology, the LSTM can directly be applied. In LSTMs the recurrent layers normally found in recurrent ANNs are replaced by purpose-built memory cells, with content controlled by different gate types (input, forget, and output). The outputs of an LSTM memory block are sent back to block inputs and gates through recurrent connections. For a more detailed description of LSTMs see~\citep{hochreiter1997long}. LSTMs have shown promise in plant classification by taking into account plant growth over time \citep{namin2017deep}, but to the best of our knowledge they have not yet been applied to learn a plant growth forward model. 

{\it Evolutionary Computation.}~~
Evolutionary approaches have been applied to many areas of robotics, including design of robot controllers~\citep{bongard13}. The approach we utilize, NEAT (NeuroEvolution of Augmenting Topologies)~\citep{stanley04}, is a an evolutionary algorithm that evolves ANNs incrementally from simple initial networks while preserving the diversity of the evolutionary population.
Evolution of robot controllers can follow an embodied approach~\citep{watson02}, meaning that the controller is evolved directly in the real hardware. Another approach is to evolve controllers in simulation based on relevant models, and then transfer the evolved controller to the real hardware. While the former approach can be time-consuming and costly, the latter can suffer from the reality gap problem~\citep{koos13}, meaning that the evolved controller performs poorly on the real hardware due to unknown limitations of the models.

\section{Methods}
Our machine learning approach to shaping natural plants follows the methods below. First, preliminary data collection experiments in the bio-hybrid setup record plant growth patterns in reaction to light stimuli. Recorded images are processed to build a stem shape dataset. This data is used to train an LSTM in a supervised way, to simulate plant stem stiffening and motion under any given sequence of the light stimuli.
Finally, the LSTM network is used as a forward model to evolve controllers in simulation, for the task of steering and shaping a plant, to avoid hitting obstacles and reach desired targets by exploiting stem stiffening phenomena.

\subsection{Bio-hybrid setup}
\label{sec:biosetup}
Following our approach in~\citep{wahby15,wahby2016evolutionary, Hofstadler2017taas}, the bio-hybrid setup is enclosed in a commercial grow box of dimensions $ 120 \times 120 \times 200 $~cm$^3$ in width, depth and height.
The grow box interior is lined in matte black board for a consistent background, diminishing light reflections.
Freshly germinated beans (`Saxa' variety\footnote{See our previous work~\citep{Hofstadler2017taas} for all product specifications in this section.}) are placed in commercial soil in 1.5l-pots, aligned with the grow box back midpoint.
The centralized robotic element consists of the following: two NeoPixel LED strips
, a~ Raspberry Pi camera module
, an LED light-bulb
, and a~Raspberry Pi.
A NeoPixel strip contains 144 RGB LEDs
, with peak-emission at wavelengths $\lambda_{\text{max}}$ 630, 530, and 475~nm respectively. Each LED can emit up to 18~lumens at full power, consuming~0.24~W.
The NeoPixel LED strips are coiled into cylindrical shapes and fastened to the grow box back wall, 30~cm above the soil and 35~cm to either side.
The camera module faces the plant at a height of 32~cm and distance of 74~cm.
The LED light-bulb is used as a flash when photographing, at 80~cm above the ground and centered over the pot.
The Raspberry Pi runs background processes\footnote{Managed by Systemd, system and service manager for Linux operating systems.} to administrate plant experiments, including synchronizing flashes, capturing photos, extracting plant stem data, running ANNs, controlling light sources, and uploading data to a Network-attached storage device (NAS).\footnote{The ZeroMQ (\url{http://zeromq.org/}) library is used for communication among these.}

\subsection{Model setup}
\subsubsection{Dataset experiments}
Our plant model is derived from our previous dataset experiments with real plants, in a bio-hybrid setup. These include six~repetitions with a simplistic, non-reactive controller
~\citep{wahby15}, and three~repetitions with a closed-loop adaptive controller
~\citep{Hofstadler2017taas}. The open-loop controller switches light sources in regular six~hour intervals, and the closed-loop controller switches according to plant tip position.
In each experiment the plant is
photographed every five~minutes. 
The plants show influence by both growth and motion, with substantial motion horizontally. They also show variance between the behaviors of individual plants.
Observing variance in plant experiments
is a well-known phenomenon in plant science, which requires high numbers
of repetitions. However, in the
context of this research, where the focus is on evolutionary
computation and robotics, such high overheads for experiments are
infeasible. Instead we test our approach based on an engineering
perspective by testing whether the model, that results from these
experiments, helps us to successfully control a plant. We also test
if the evolved controllers are able to perform properly
with such dynamic and unexpected plant behavior.

\subsubsection{Stem motion tracking}
\label{section:stem_motion_tracking}
We describe our computer vision method for stem motion tracking. The 10-point description of stem geometry forms the basis of training data for our LSTM-based {\it Stem stiffening and motion model}. 
We process images from the dataset experiments described above, to record a 10-point $xy$ description of the full stem at each timestep, representing its phototropic motion and stiffening dynamics. 10 points is sufficient to capture curvature details within the growth area of the current setup. The images are sampled\footnote{\label{fn:sampling}Sampling was duplicated in two platforms: Python, utilizing the OpenCV library; and IronPython using Grasshopper libraries pertaining to computer vision.} at 1/8 resolution and processed according to the following method, both for the dataset experiments described above, and for the reality gap experiments detailed in Sec.~\ref{sec:results:reality}.
Before processing the dataset experiment images, a set of images is compiled showing the setup without a plant. The setup images include all states of the controller and any slight variations in lighting conditions. The set of images is sampled,\textsuperscript{\ref{fn:sampling}} isolating
the green RGB channel value at each pixel position~$(i,j)$ and remapping it onto the domain~$[0,1]$, forming sequence~$\Lambda$ containing a matrix~$M$ of green values for each image. To represent the interval of possible green channel values present in the setup, matrices~$L$~and~$H$ are constructed by
\begin{equation}
\begin{aligned}
	& L_{i,j} = \min_{M \in \Lambda} (M_{i,j}), 
	& H_{i,j} = \max_{M \in \Lambda} (M_{i,j}).
\end{aligned}
\end{equation}

After constructing the setup matrices, dataset experiment images of plants are processed. The green channel value is isolated for each pixel~$(i,j)$, remapped to the domain~$[0,1]$, and saved into the matrix~$R$. Pixels within a window are identified as containing plant material if: $(R_{i,j} < L_{i,j} - \theta_1) \lor (R_{i,j} > H_{i,j} + \theta_1)$, for threshold \mbox{$\theta_1 = 0.2$}. Each identified plant pixel is extracted to set~$P$, and their $(x_p,y_p)$ coordinate positions are used to identify two possible locations of the plant's tip. In order to locate the growth tip~$g = (x^g,y^g)$, plant pixels are compared to the globally defined anchor~$a = (x^a,y^a)$, representing the position where the plant stem emerges from the soil. Two possible $xy$ growth tip positions are identified (corner point~$c$ as the furthest from $a$ in $xy$, and high point~$h$ as the same in $y$ only) and one selected as $g_n$ based on Euclidean distance $(d)$ from $g$ in the prior timestep, such that
\begin{align}
c = \argmax_{(x_p,y_p) \in P} \vert x^a - x_p \vert + \vert y^a - y_p \vert, \\
h = \argmax_{(x_p,y_p) \in P} \vert y^a - y_p \vert, \\
g_{n}=\left\{
  \begin{array}{@{}ll@{}}
    h, & \text{if}\ d(g_{n-1},h) < d(g_{n-1},c) \\
    c, & \text{otherwise}
  \end{array}.\right.
\end{align}

The remaining intermediate points $((x^{S_{\it 2}},y^{S_{\it 2}}), \cdots , (x^{S_{\it 9}},y^{S_{\it 9}}))$ describing the stem are preliminarily identified from set $P$, and then smoothed while preserving the representation of stiffening dynamics. For these eight points, the $y^{i}$ are distributed evenly between the tip and anchor, as
\begin{equation}
y^{S_i} = \frac{i}{9} \vert y^a - y^g \vert + y^a , \\
\end{equation}
and $x^{S_i}$ are set as the averaged $x_p$ for pixels in set $P$ that have $y_p$ within threshold $\theta_2$ of the respective $y^{S_i}$, such that 
\begin{equation}
x^{S_i} = \overline{x_p} : \forall x_p \in P : \vert y_p - y^{S_i} \vert < \theta_2 , \
\end{equation}
where $\theta_2 = 30$ pixels. In this way, a 10-point description $S$ of the full stem $S = ( a, (x^{S_{\it 2}},y^{S_{\it 2}}), \cdots , (x^{S_{\it 9}},y^{S_{\it 9}}), g)$ is defined. Due to minor variations in images caused by shadows and light reflections on the stem, this 10-point detection contains some errors. We address these errors using a simple algorithm based on Smoothing via Iterative Averaging (SIA) \citep{mansouryar2012smoothing}, which preserves the key topological features of the curve being smoothed. For each point in $S_n$, our algorithm utilizes the equation
\begin{equation}\label{SIA}
(x^{S_i},y^{S_i}) = \left(\frac{1}{2}(x^{i-1} + x^{i+1}),\frac{1}{2}(y^{i-1} + y^{i+1})\right)
\end{equation}
iteratively, according to the following steps: 1) for $i \in \{2, 4, 6, 8\}$, apply eq.~\ref{SIA}, 2) for $i \in \{3, 5, 7, 9\}$, apply eq.~\ref{SIA}, 3) for $i \in \{2, 4, 6, 8\}$, apply eq.~\ref{SIA}.
In this way, the intermediate stem points are smoothed with the SIA-based process, while the tip and anchor remain unchanged. The newly smoothed sequences $S$ are converted to cm and scaled to match physical setup dimensions. The anchors are then unified to standardize the data, by translating $(x^{S_i},y^{S_i})$ points in $S_n$ according to 
$((x^{S_i},y^{S_i}) + (a - \overline{a} : \forall a \in S_n))$.
The resulting data is reformatted to sequence $\Psi$ of $18$-dimensional vectors 
$
\psi_j = (x^{S_{\it 2}}_j,y^{S_{\it 2}}_j, \cdots ,x^{S_{\it 9}}_j,y^{S_{\it 9}}_j,x^g_j,y^g_j),
$
without the now redundant anchor values. These vectors are the basis for regression data for our LSTM-based {\it Stem stiffening and motion model}, described below.

\subsubsection{LSTM trained as Stem stiffening and motion model}
\label{section:stem_motion_model}

Building a holistic model of plant stem dynamics is a complex task~\citep{bastien2015unified} that would benefit from deep learning. However there is a lack of existing data, and the substantial overhead associated with plant experiments makes it infeasible to obtain large amounts of new data (many plants can be grown in parallel but controlled light conditions, monitoring, and tracking are costly). Therefore, having obtained a small amount of data from real plant experiments---described above---we develop a method to artificially expand that data, avoiding overfitting when training the LSTM. 

\paragraph{Preparation of stem data for regression}
After manually removing data in $xy$-areas that are too sparsely populated to provide reliable data (mostly in zones far from the origin, where only one plant of nine may have reached by coincidence), we process the experiment motion tracking data in two ways to expand the set. Firstly we add noise, to reduce the tendency of overfitting, and secondly we add a generic model, such that the typical plant behavior is
dominantly represented in the data for
the LSTM.

In order to add normal distribution noise, in addition to the experiment data in sequence $\Psi$, we define noisy data in sequences $(\Psi_1,\cdots,\Psi_n)$, where
\begin{align}
   \psi_{n_j} = ((x^{S_{\it 2}}_j)^{\Psi_n}, (y^{S_{\it 2}}_j)^{\Psi_n}, \cdots ,(x^{S_{\it 9}}_j)^{\Psi_n},(y^{S_{\it 9}}_j)^{\Psi_n}, \linebreak (x^g_j)^{\Psi_n},(y^g_j)^{\Psi_n}).
\end{align}
The noise values applied to each growth tip $(x^g_j,y^g_j)$ in $\Psi$ are computed according to the mean $\mu$ and standard deviation $\sigma$ of a finite quantity ($\theta_3$) of the closest growth tips $(x^g_i,y^g_i)$ that have the same light condition $b$. To calculate this, for each growth tip $(x^g_j,y^g_j)$, all other growth tips in $\Psi$ sharing the same light condition $b$ are placed into sequence $\mathbf{dist}^j$, and are then sorted according to their Euclidean distance from the respective tip at $m = j$. The $\theta_3$ closest tips for each respective tip are defined as $W_{\Psi}$, such that 
\begin{align}
    w_j = (x^g_i,y^g_i) \in \Psi \quad | \quad  n \leq \theta_3 \in \mathbf{dist}^j_n,
\end{align}
where $\theta_3 = 100$. 
The mean $\mu$ for noise is calculated as 
\begin{align}
    \mu(x^g_j) = \frac{1}{|W_{\Psi}|} \sum_{j=1}^{|W_{\Psi}|}{w_j x^g_i},
\end{align}
with a symmetrical equation for $\mu(y^g_j)$, and standard deviation $\sigma$ as
\begin{align}
    \sigma^2(x^g_j) = \frac{1}{|W_{\Psi}|} \sum_{j=1}^{|W_{\Psi}|}{w_j \left(x^g_i - \mu(x^g_j)^2 \right)},
\end{align}
with a symmetrical equation for $\sigma^2(y^g_j)$.
The noisy data in each new sequence $\Psi_n$ is calculated by first defining the noisy growth tips and then defining the noisy intermediate points in relation to the tip output. The new noisy tips $((x^g_j)^{\Psi_n},(y^g_j)^{\Psi_n})$ in $\Psi_n$ are calculated using normal distribution noise, according to the existing tips $(x^g_j,y^g_j)$, and $\mu$ and $\sigma$ values scaled by factor $\omega$ such that
\begin{align}
(x^g_j)^{\Psi_n} = x^g_j + \mathcal{N}\left(\mu (x^g_j), \sigma (x^g_j) \omega\right),
\end{align}
with a symmetrical equation for $(y^g_j)^{\Psi_n}$, where scaling factor \mbox{$\omega = 0.1$}. 
The noisy intermediate points $((x^{S_i}_j)^{\Psi_n},(y^{S_i}_j)^{\Psi_n})$ in $\Psi_n$, are calculated according to the noisy tips $((x^g_j)^{\Psi_n},(y^g_j)^{\Psi_n})$, and are scaled by the values $\omega_2(x^{S_i})$, $\omega_2(y^{S_i})$. The noise values are generated through an artificial mean $\mu_2$ and standard deviation $\sigma_2$, defined according to the calculated standard deviation and the generated change in position of the noisy growth tips, such that
\begin{align}
\mu_2\left((x^{S_i}_j)^{\Psi_n}\right) = x^{S_i}_j + \left((x^g_j)^{\Psi_n} - x^g_j\right) \omega_2(x^{S_i}),\\
\sigma_2\left((x^{S_i}_j)^{\Psi_n}\right) = \left(\sigma (x^g_j) \omega\right) \omega_2(x^{S_i}), 
\end{align}
with symmetrical equations for $\mu_2\left((y^{S_i}_j)^{\Psi_n}\right)$, $\sigma_2\left((y^{S_i}_j)^{\Psi_n}\right)$, where scaling factors $\omega_2(x^{S_i})$, $\omega_2(y^{S_i})$ are defined according to the extents in $\Psi$ of the respective intermediate point $(x^{S_i},y^{S_i})$ in comparison to the extents of growth tip $(x^g,y^g)$. These are defined as
\begin{align}
\omega_2(x^{S_i}) = |\min_{x^{S_i}_i \in \Psi} (x^{S_i}_i) - \max_{x^{S_i}_i \in \Psi} (x^{S_i}_i) | \cdot
|\min_{x^g_i \in \Psi} (x^g_i) - \max_{x^g_i \in \Psi} (x^g_i) |^{-1},
\end{align}
with a symmetrical equation for $\omega_2(y^{S_i})$.
In new noisy data $\Psi_n$, the intermediate points $((x^{S_i}_j)^{\Psi_n},(y^{S_i}_j)^{\Psi_n})$ are defined using normal distribution noise, according to $\mu_2$ and $\sigma_2$ and scaled by $\omega$, such that
\begin{align}
(x^{S_i}_j)^{\Psi_n} = x^{S_i}_j + \mathcal{N}\left(\mu_2((x^{S_i}_j)^{\Psi_n}),\sigma_2((x^{S_i}_j)^{\Psi_n}) \omega\right),
\end{align}
with a symmetrical equation for $(y^{S_i}_j)^{\Psi_n}$.
In the methods implementation described in this paper, we conduct three runs of equations~11-15 and their respective symmetries, generating three unique sequences of noisy data: $\Psi_1, \Psi_2, \Psi_3$.

In order to add a generic plant model, we manually select experiment data associated with the natural plants' smoothest and least noisy movements (identified by observation), and then follow a data-driven approach. We reinforce these generic movements as dominant by adding additional translations of them, distributed over small $xy$ distances. In addition to experiment data $\Psi$ and noisy data $\Psi_n$, we define new generic model data in sequence $\Psi_\Phi$, with each vector $\psi_{\Phi_j}$ structured as those in $\Psi$, defined as:
\begin{align}
(x^{\Psi_\Phi}_j,y^{\Psi_\Phi}_j) \in \psi_{\Phi_j} = \left( (x_j \in \psi_j) \pm 10^\lambda, (y_j \in \psi_j) \pm 10^\lambda \right),
\end{align}
where $\lambda = (-3,\dots,-6)$, generating $64$ new $xy$ translations in $\Psi_\Phi$.

The data sequences $\Psi, \Psi_n,$ and $\Psi_\Phi$ are combined to form $\mathbf{\Psi}^{*}.$ Vectors in $\mathbf{\Psi}^{*}$ are then mirrored across the $x$-axis, as we assume the targeted plant behavior to lack left-right bias. This also doubles the quantity of data. Then $\mathbf{\Psi}^{*}$ is reformatted according to timestep, such that each new vector $\mathbf{\psi}^{*}_j$ is composed of 18 dimensions representing the current $xy$ stem position, 18 dimensions representing the next stem position, and one dimension representing the Boolean light condition. Vectors are removed if they 1) contain duplicate stems at the current and next positions, or 2) if the $xy$ change is greater than $20\times$ the average $xy$ in $\mathbf{\Psi}^{*}$. We end up with a $\mathbf{\Psi}^{*}$ dataset containing 101,162~vectors.

\begin{figure}
\centering
\hspace{-0.4in}\includegraphics[width=2.5in]{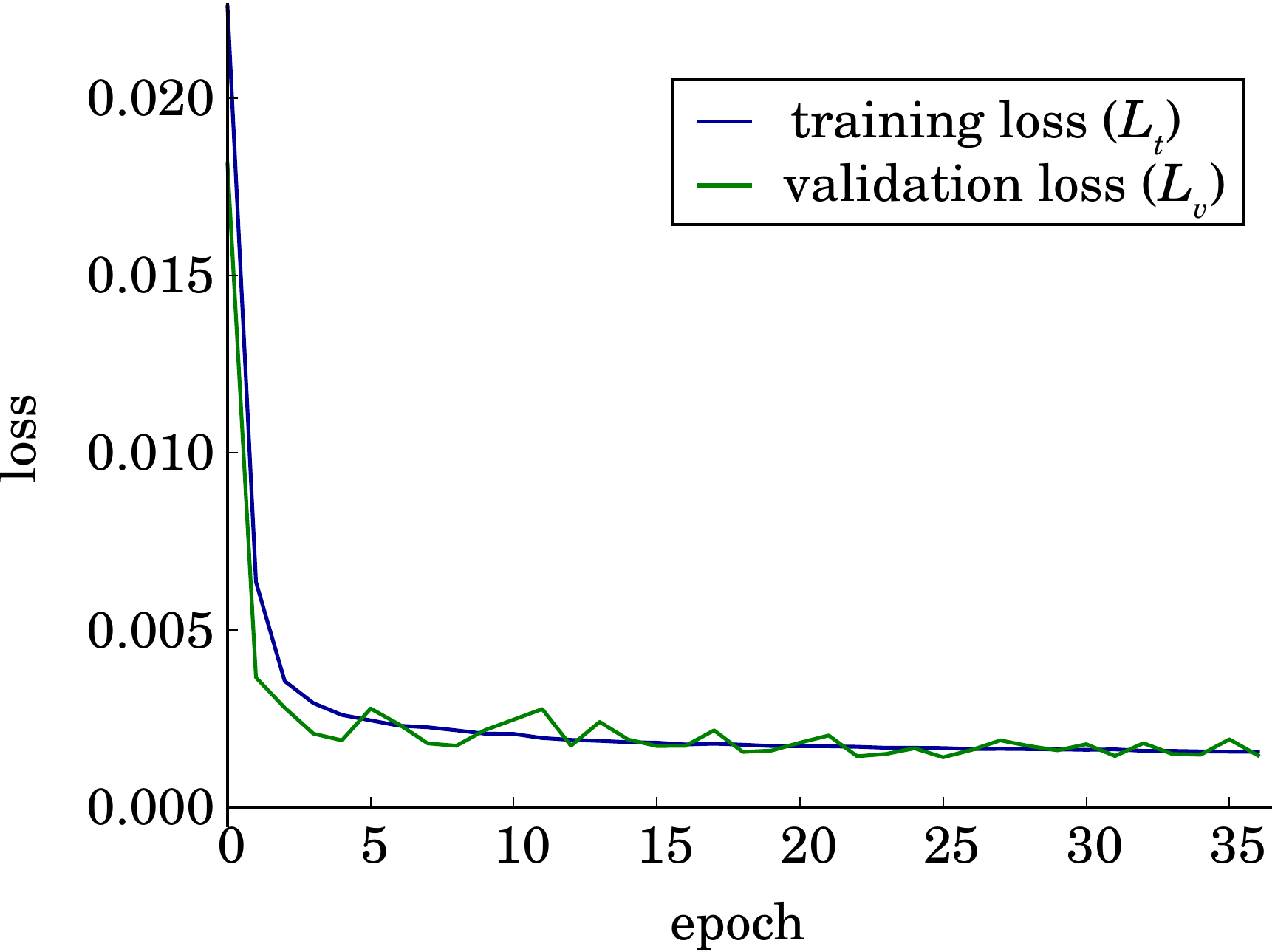}
\vspace{-0.1in}
\caption{LSTM-based model training.}
\label{fig:LSTM}
\end{figure}

\paragraph{Training procedure}
In order to obtain a holistic plant model, 
we train the
LSTM using Keras~\citep{chollet2015keras}, a high-level wrapper of TensorFlow~\citep{abadi2016tensorflow}.
The data in $\mathbf{\Psi}^{*}$ is formatted as described above, in vectors containing nine 2D stem points at a given time step and at the subsequent timestep, together with current light conditions (left/right light on). The LSTM network has 19~input units (current nine points and light condition), 50~LSTM memory blocks, and 18~output units (next nine points).
We shuffle the vectors $\mathbf{\psi}^{*}_j$ and split them into training (70\%), validation (20\%), and testing set (10\%). 
We train the LSTM network with the training set in sequence of batches of size $N = 30$ for 200 epochs at a steady learning rate of 0.001 using Adam optimizer~\citep{abadi2016tensorflow}.
The training loss $L_t$ is the mean absolute error (MAE), defined as
\begin{align}
L_t = & \frac{1}{N} \sum _{i=1}^{N} \frac{1}{18} \sum _{j=2}^{10} \mid x^{j}_p - x^{j}_t \mid + \mid y^{j}_p - y^{j}_t \mid ,
\label{eq:FF}
\end{align}
where $x^{j}_t$ and $y^{j}_t$ are the true $xy$ coordinate values of the stem point~$j$, and $x^{j}_p$ and $y^{j}_p$ are the corresponding predicted coordinates.
The validation dataset is used to track the training progress through validation loss $L_v$ (calculated similarly to $L_t$ but not in batches). An early stopping callback is implemented to prevent overfitting by tracking $L_v$ and stop the training process with patience of ten epochs (i.e., if the $L_v$ stops improving for ten epochs).
As seen in Fig.~\ref{fig:LSTM}, the training process stops at the 27th epoch when $L_v$ stopped improving for ten epochs at $L_t = 1.56\times 10^{-3}$ and $L_v = 1.55\times 10^{-3}$.
Then, we calculate the MAE for each of the three datasets when used as input to the network. The error values for the training, validation and testing datasets 
are $1.55\times 10^{-3}$, $1.55\times 10^{-3}$, and $1.44\times 10^{-3}$ respectively.
On average, the error is $\approx 1$~mm at each coordinate value, showing that the resulting model represents plant behavior closely\footnote{Find a video at: \url{https://vimeo.com/265144652}}.

\subsection{Controller setup}

Our controller is an ANN operating two light sources.
The input to the ANN at each time step is 1) a set of 10-points (20~real numbers) representing the current plant position and shape, as described above, 2) the current coordinates of the target (2~real numbers), and 3) coordinates of the obstacle (4~real numbers).
We have two setups: in silico (simulation) and in vivo (`wet' setup with plant and hardware). In silico, the $10$ points are directly generated using the \emph{stem stiffening and motion model}. In vivo, a camera and computer vision detects the actual plant and forms the corresponding $10$ points. The output of the ANN is the control triggering light sources for stimuli. 

\subsubsection{Task: Obstacle avoidance by shaping the plant}
\label{sec:task}

The controller has to shape the plant appropriately by navigating it around a virtual obstacle to then reach a target area (radius is 2 cm).
The plant should not touch the obstacle with any part of its body throughout the experiment.
Since the obstacle is virtual, it neither casts a shadow, nor does it give other physical cues (e.g., a mechanical barrier) that would allow the plant to avoid or grow around it by itself.
We perform the obstacle avoidance task in two different experiment settings.
In the first experiment (\textit{left target} experiment), a fixed target is located at 5.12~cm to the left of the plant anchor and 17.9~cm above it.
We evaluate the controller in four different scenarios where a rectangular obstacle ($7 \times 3$~cm$^2$) is centered at four different locations.
In the first scenario the obstacle is centered ${\approx}8.24$~cm left of the plant anchor point and at a height of $8.8$~cm. In the second scenario, the obstacle is 2.67~cm further to the right (closer to the plant), making the task more challenging. In the third scenario, the obstacle is an additional 2.67~cm further to the right, making it impossible for the plant to reach the target. Finally, the obstacle is an additional 5.33~cm further to the right, this time clearing the area enough for the plant to again reach the target.
In the second experiment (\textit{middle target} experiment), a fixed target is located above the plant anchor at a hight of 17.9~cm. Here, we have only two scenarios. In the first scenario, the obstacle is centered at ${\approx}3$~cm right of the plant anchor and at a height of $8.8$~cm. In the second scenario, the obstacle is centered at ${\approx}3$~cm left of the plant anchor and at a height of $8.8$~cm.
Hence, the controller requires different strategies to control the plant for different target/obstacle configurations, which makes the task more challenging.
In addition, the plant stiffens only over time, requiring the plant tip to be guided in wide deviations from the plant's ending configuration.

\subsubsection{Evolutionary approach}

We use MultiNEAT~\citep{multineat}, a portable library that provides Python bindings to NEAT~\citep{stanley04}, to evolve ANN controllers.
We use the NEAT parameters set from 
our previous work~\citep{wahby15, wahby2016evolutionary, Hofstadler2017taas}.
We follow a step-wise simulation approach, where the stem description $S_t = (x^a_t,y^a_t,x^{S_{\it 2}}_t,y^{S_{\it 2}}_t, \cdots ,x^{S_{\it 9}}_t,y^{S_{\it 9}}_t,x^g_t,y^g_t)$, the target position ${\bf x}_{i}^{*}$,
and the coordinates of an obstacle ${\bf x}_{i}^{o}$ are input to the ANN at each time step $t$. The output of the network~($C_t$) regulates
the light settings. If $C_t \leq 0.5$, it triggers the left light source, otherwise, the right.
The current plant condition and light setting $({\bf x},C)_t$ impact plant behavior during that time step. The \textit{LSTM stem stiffening and motion model} is used to predict the next plant stem $S_{t+1}$ accordingly. 
For experiments in reality, an image of the plant is processed to determine $S_{t+1}$ (see Sec.~\ref{section:stem_motion_tracking}).
The simulation is stopped when the tip $y^g_t$ value is equivalent to ${\approx}21$ cm or once the plant touches an obstacle.
Beans require ${\approx}72$ hours to grow that high. This overhead is relatively manageable, and
allows enough growth to exploit stem stiffening and avoid obstacles.

At the simulation end (at $t = f$), performance of the ANN controller is evaluated using a behavioral fitness function~$F$ (according to the classification in~\citep{nelson_2009_fitness}).
Plant motion is rewarded by measuring the distance traveled by tip $g$ towards the target along both $x$ and $y$~axes as 
 $x_r = | x^{*} | - | x^{*} - x^g_{f} |$, 
 $y_r = | y^{*} | - | y^{*} - y^g_f |$,
where~$(x^{*},y^{*})$ is the target position.
The fitness $F$ is then calculated by 
\begin{align}
F = &\hspace{1.5 mm} \frac{x_r + y_r}{| x^{*} | + | y^{*} |},
\label{eq:delta}
\end{align}
where $| x^{*} | + | y^{*} |$ is the theoretical best fitness value the controller can achieve.
If the controller is evaluated at different scenarios, then its fitness value is the average of all evaluations.

\section{Results}

Based on the \emph{stem stiffening and motion model} (see Sec.~\ref{section:stem_motion_model}), we evolve the robotic controllers and evaluate their performance in simulation. Next, we transfer the fittest controllers to reality and investigate the extent of the reality gap.

\subsection{Evolving controllers in simulation}
\label{subsec:sim_results}

\begin{figure*}
   \centering
   \subfigure[\label{fig:results:sim:box:1}\textit{left target} exp., boxplot of best fitness per gen.]{
     \includegraphics[angle=0,width=2.2in]{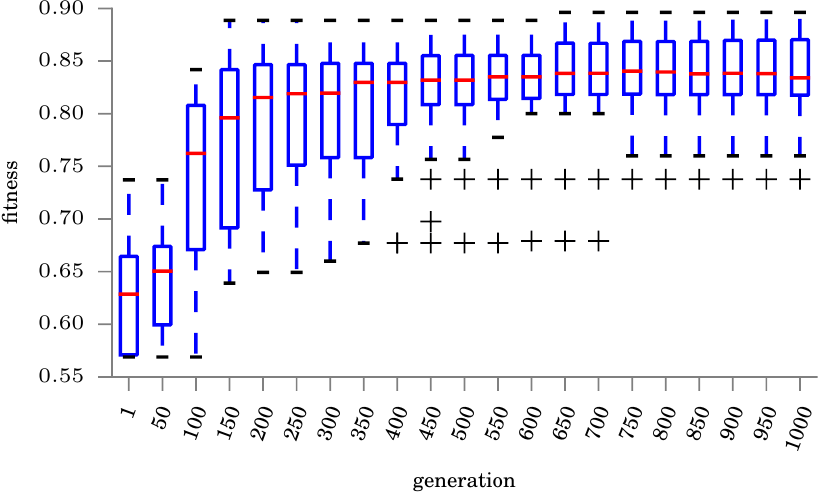}
   }
   \hspace{2mm}
   \subfigure[\label{fig:results:sim:fbox:1}\textit{left target} exp., functional boxplot, best fit. per gen.]{
     \includegraphics[angle=0,width=2.2in]{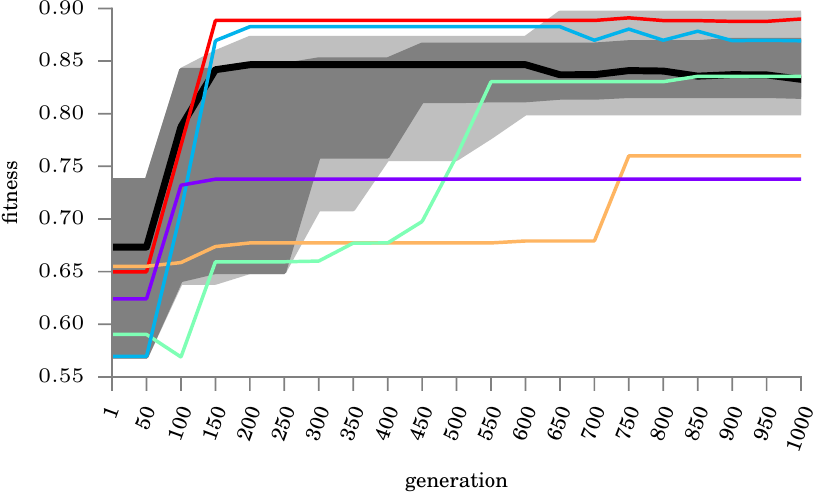}
   }\\
      \subfigure[\label{fig:results:plant:1:1}Initial stem geometry.]{
     \includegraphics[angle=0,width=1.75in]{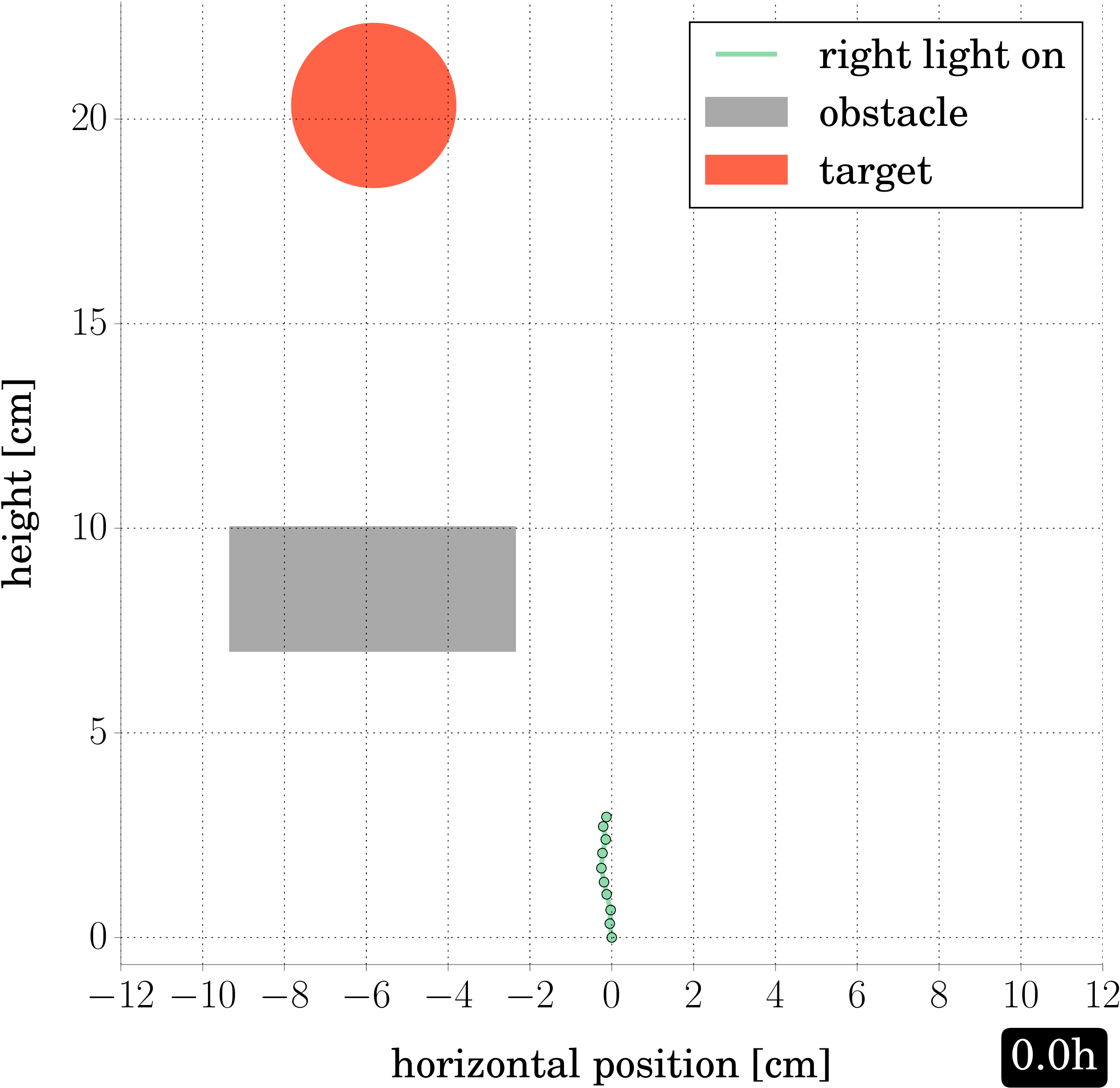}
   }
   \subfigure[\label{fig:results:plant:1:2}Stem geometry at 4.0 simulated hours.]{
     \includegraphics[angle=0,width=1.75in]{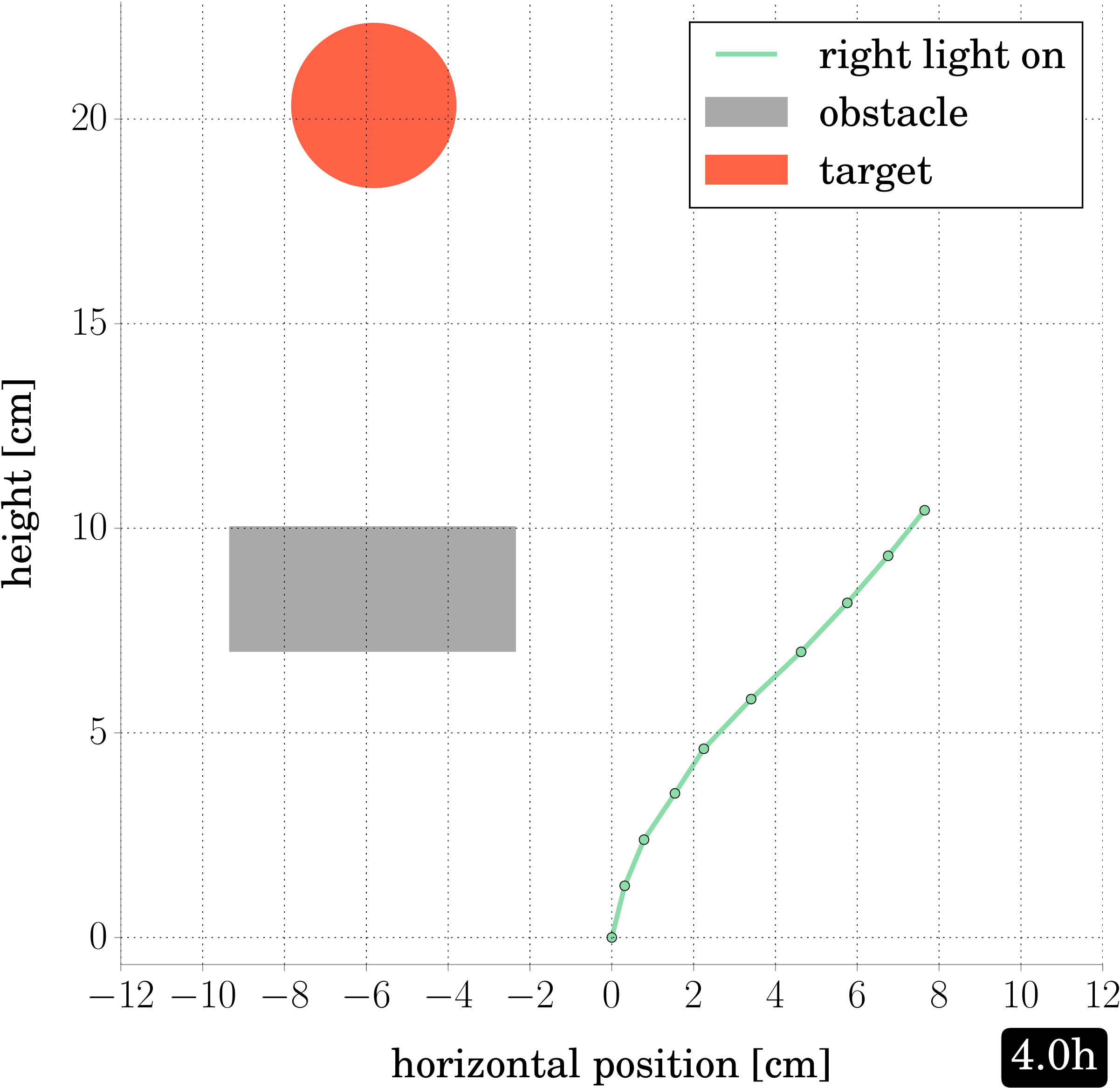}
   }
      \subfigure[\label{fig:results:plant:1:3}Stem geometry at 6.5 simulated hours.]{
     \includegraphics[angle=0,width=1.75in]{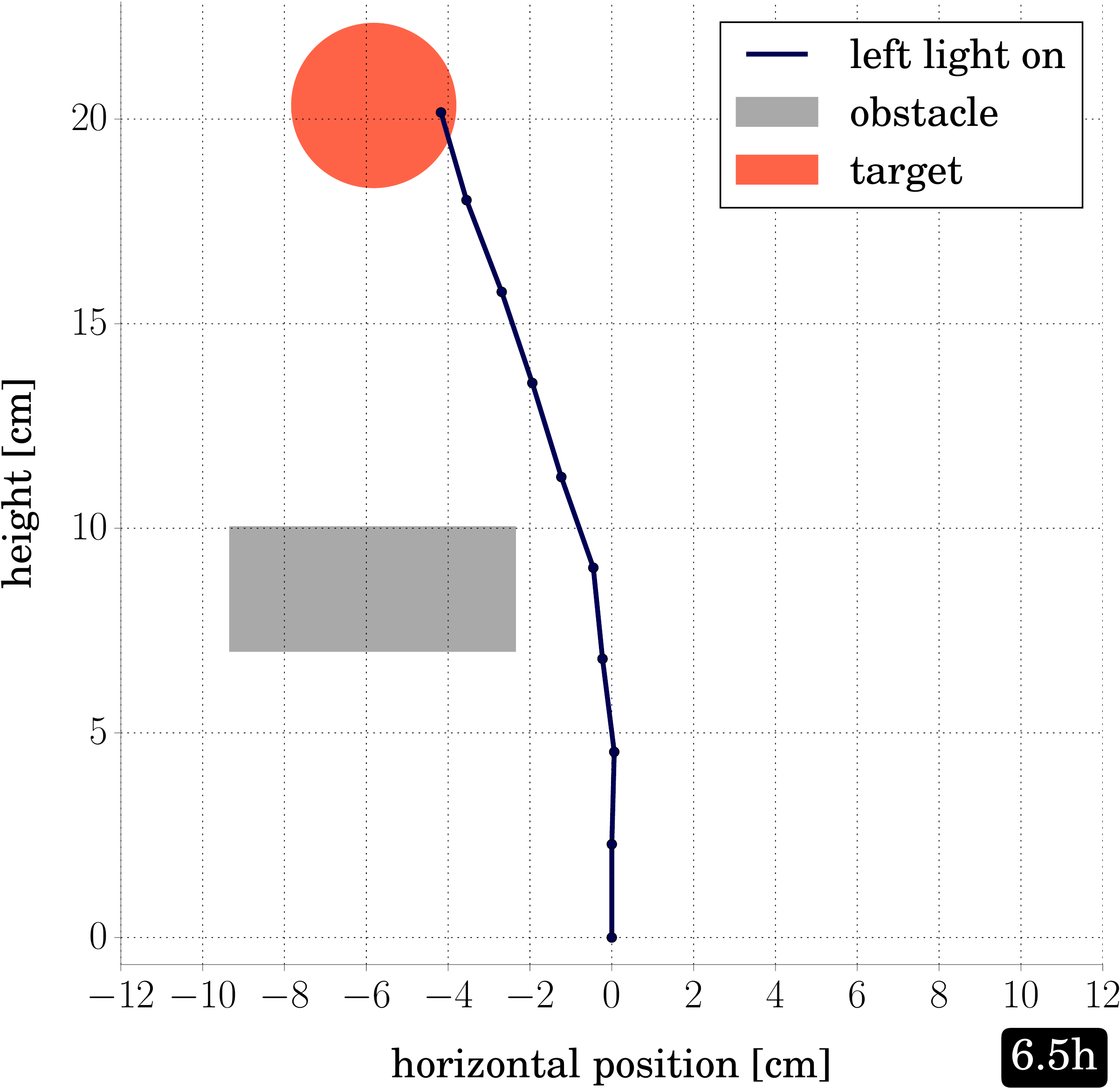}
   }
   \subfigure[\label{fig:results:sim:box:2}\textit{middle target} exp., boxplot, best fit. per gen.]{
     \includegraphics[angle=0,width=2.2in]{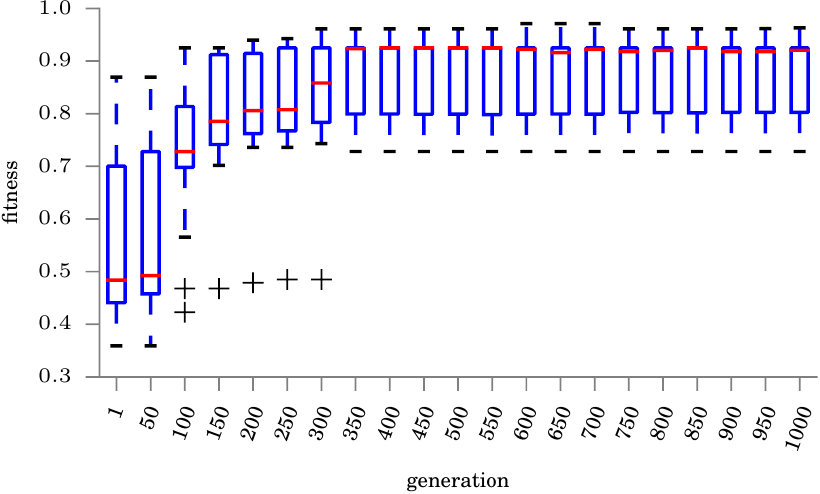}
   }
   \hspace{2mm}
   \subfigure[\label{fig:results:sim:fbox:2}\textit{middle target} exp., functional boxplot, best fit. per gen.]{
     \includegraphics[angle=0,width=2.2in]{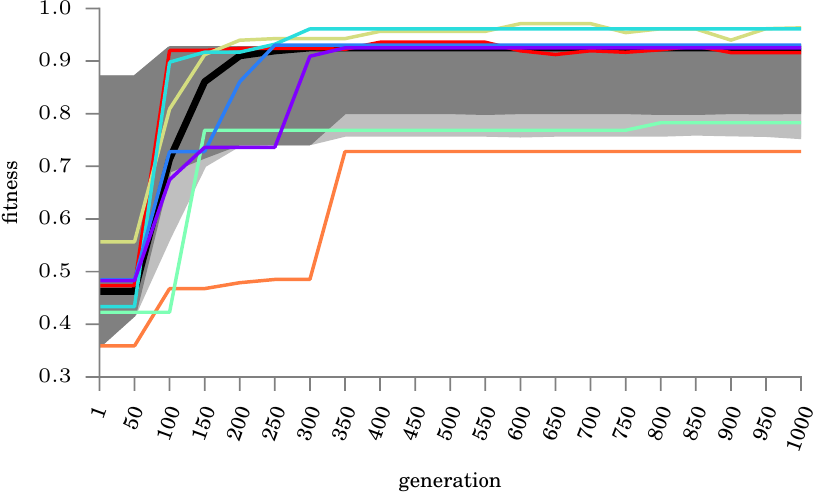}
   }\\
   \subfigure[\label{fig:results:plant:2:1}Stem geometry at 2.5 simulated hours.]{
     \includegraphics[angle=0,width=1.75in]{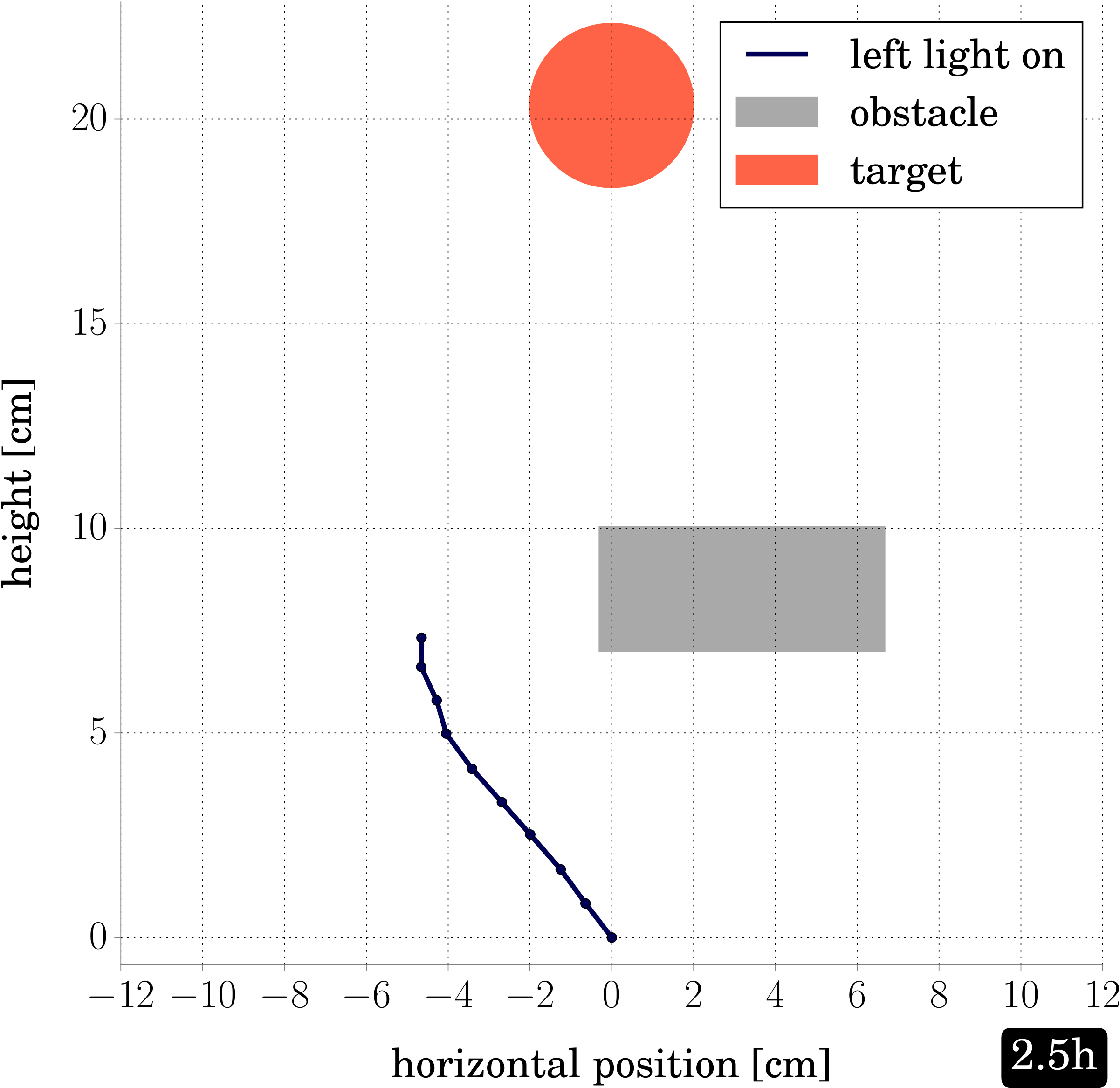}
   }
   \subfigure[\label{fig:results:plant:2:2}Stem geometry at 4.8 simulated hours.]{
     \includegraphics[angle=0,width=1.75in]{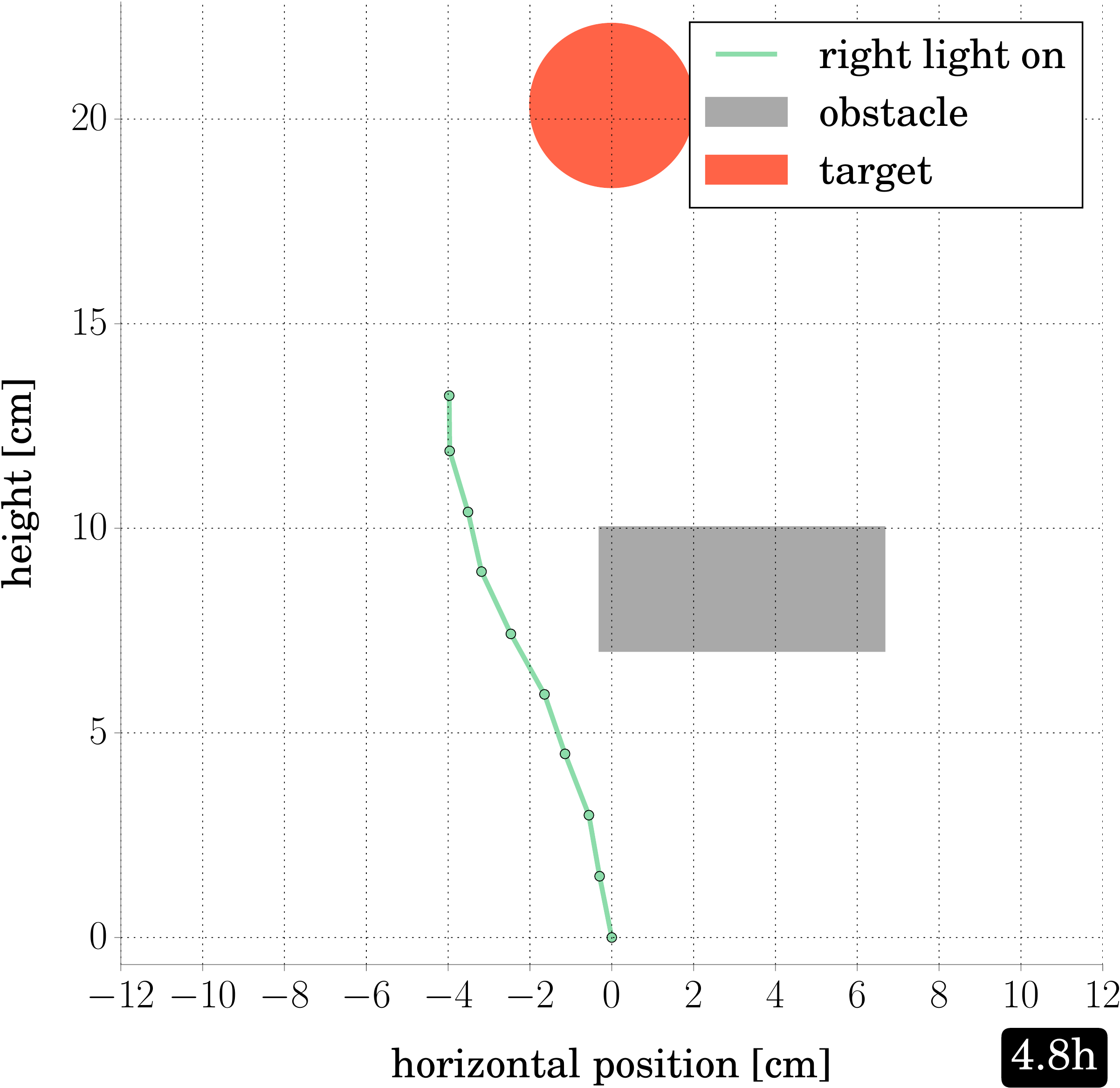}
   }
      \subfigure[\label{fig:results:plant:2:3}Stem geometry at 7.2 simulated hours.]{
     \includegraphics[angle=0,width=1.75in]{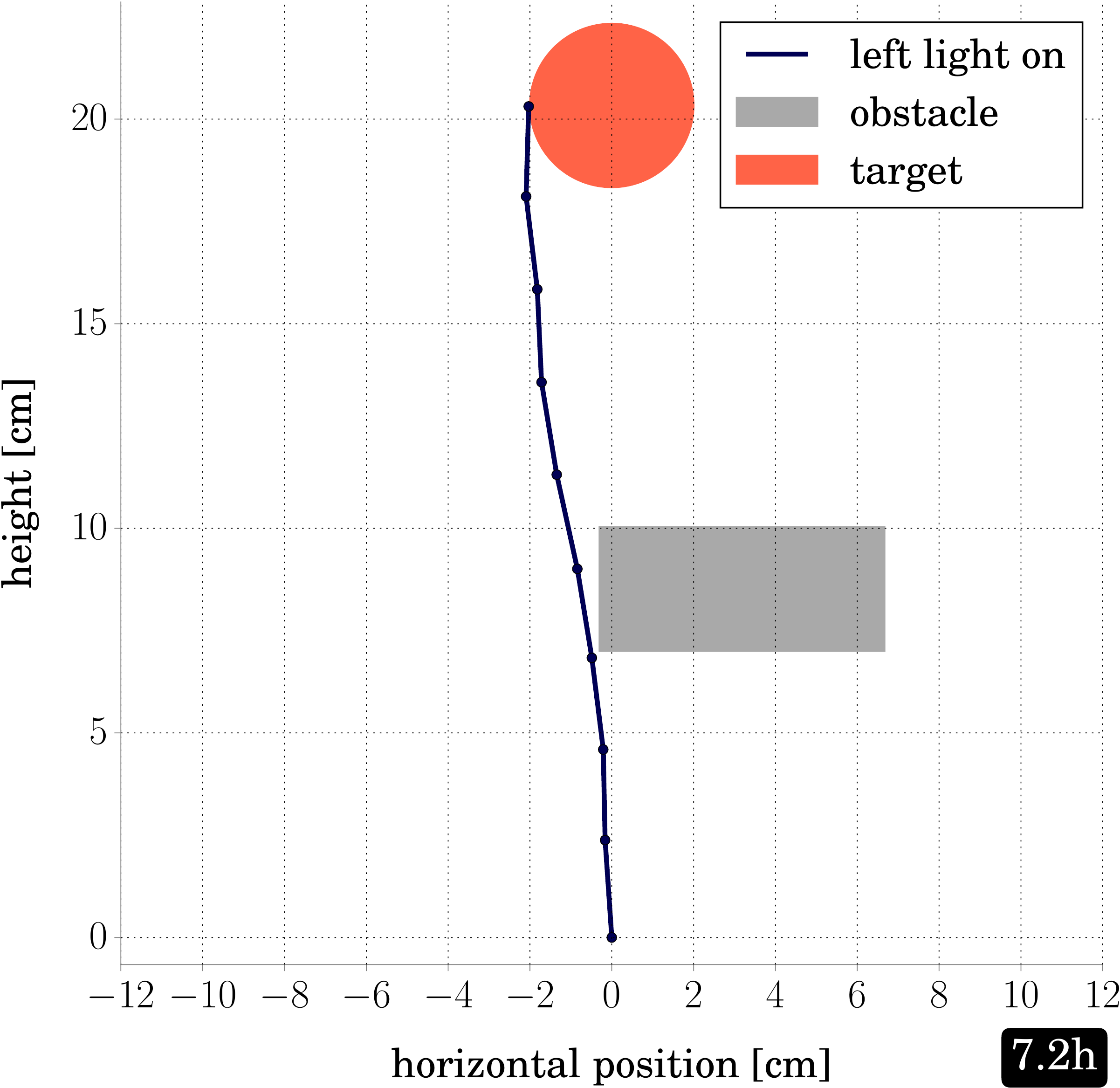}
   }
   \caption{\label{fig:results:sim:setting1} Performance of the evolutionary process over generations for 20 evolutionary runs.}
\end{figure*}

First, we report the results of the \textit{left target} experiment.
The boxplots in Fig.~\ref{fig:results:sim:box:1} and function boxplots in Fig.~\ref{fig:results:sim:fbox:1} show
the performance of 20 independent evolutionary runs, 1000 generations each.
The best fitness per generation for all evolutionary runs is considered.
Notice the steady increase in median until convergence is reached around the 500th generation.
In this experiment the controller is evaluated according to four scenarios (see Sec.~\ref{sec:task}).
According to the behavior of one of the best controllers (fitness of 82.5\%), the controller is able to determine whether or not it needs to exploit the stem's natural stiffening over time, in order to avoid hitting the obstacle.
In case there is a possibility to hit an obstacle (e.g., second scenario), the controller steers the plant away into the opposite direction of the obstacle, long enough to obtain sufficient stiffness at the lower parts of the stem, see Fig.~\ref{fig:results:plant:1:2}, then steers the plant back towards the target area, see Fig.~\ref{fig:results:plant:1:3}.
In case the obstacle is not blocking the way (e.g., forth scenario), the controller
leads the plant directly towards the target area (i.e., no stiffening is necessary).
The behavior in all scenarios can be seen in the video\footnote{\label{note:video}Find a video at:~\url{https://vimeo.com/265144652}}.

Second, we report the results of the \textit{middle target} experiment. Here, we also have 20 evolutionary runs, 1000 generations each as shown in Figs.~\ref{fig:results:sim:box:2} and~(g). In contrast to the previous experiment, the convergence is reached earlier, around the 350th generation. This indicates that the task here is easier by comparison. 
The expected behavior of the evolved controller is to first steer the plant away from the target while the stem stiffens, then steer the plant back to reach the target while avoiding the obstacle---as in the \textit{left target} experiment.
However, the controller here (fitness of 97.3\%) steers the plant to the obstacle side near the target, see Fig.~\ref{fig:results:plant:2:1}, then switches between the two light sources until
the plant obtains enough height and stiffness without hitting the obstacle, see Fig.~\ref{fig:results:plant:2:2}. Finally it steers the plant tip towards the target, see Fig.~\ref{fig:results:plant:2:3}.

\subsection{Performance of controllers in reality}
\label{sec:results:reality}

\begin{figure*}
   \centering
   \subfigure[\label{fig:results:reality:1}20 h]{
     \includegraphics[angle=0,width=0.21\textwidth] 
     {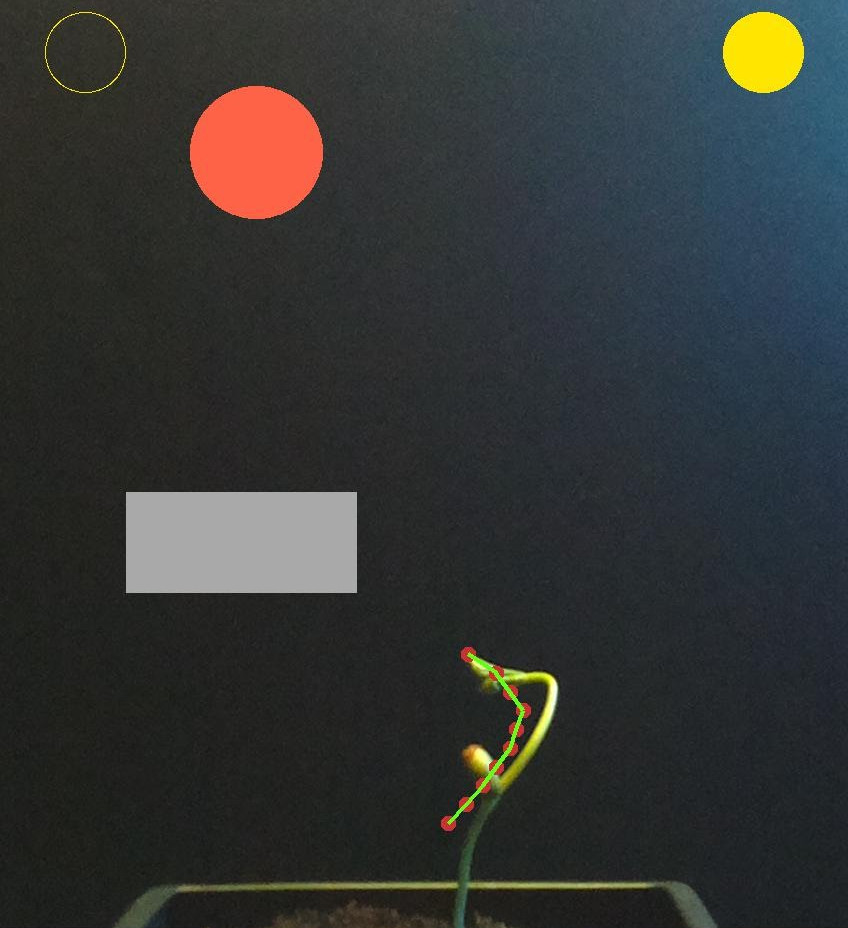}
   }
   \subfigure[\label{fig:results:reality:2}37 h]{
     \includegraphics[angle=0,width=0.21\textwidth]{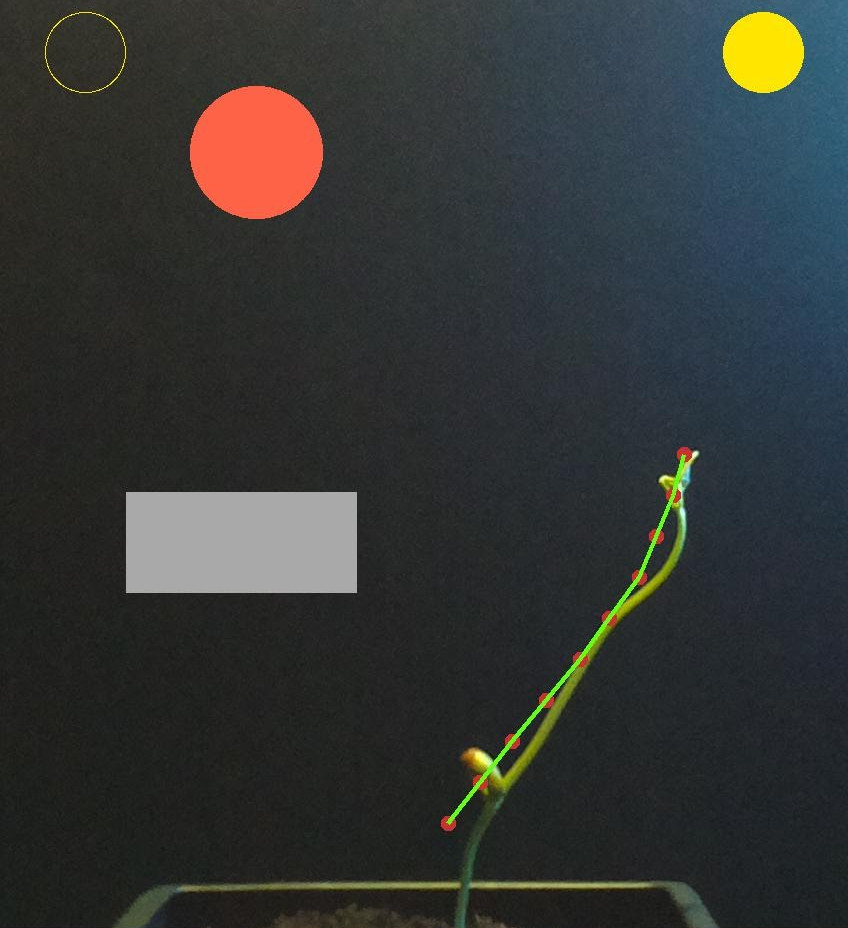}
   }
   \subfigure[\label{fig:results:reality:3}54 h]{
     \includegraphics[angle=0,width=0.21\textwidth]{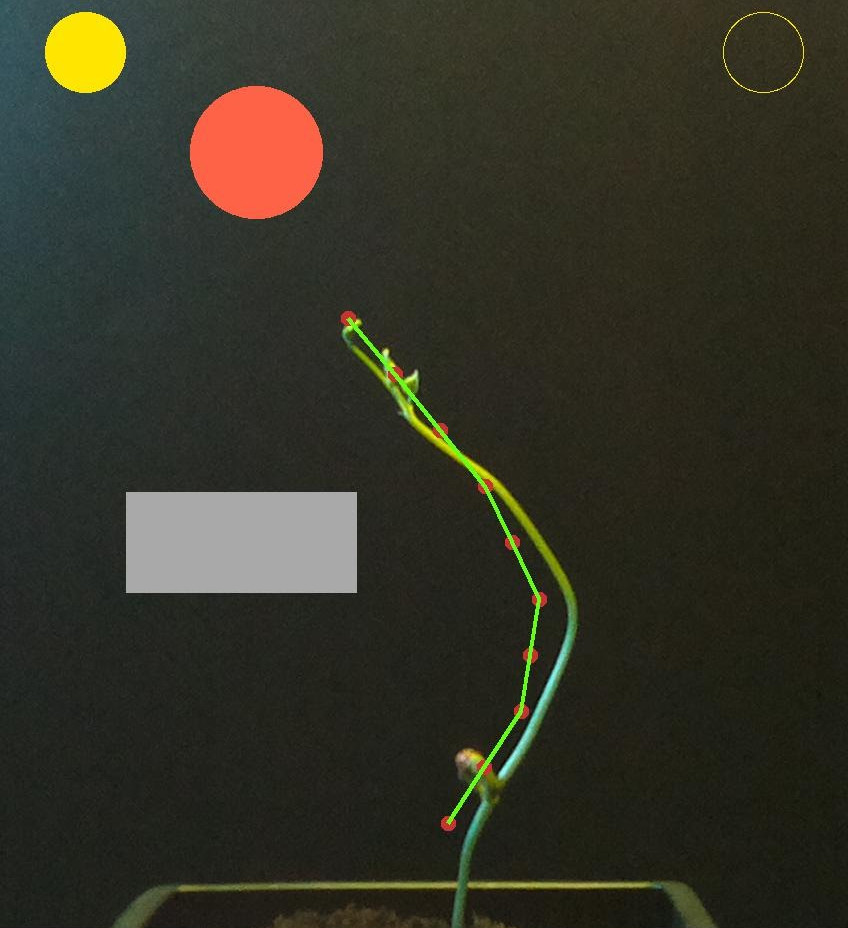}
   }
   \subfigure[\label{fig:results:reality:4}71 h]{
     \includegraphics[angle=0,width=0.21\textwidth]{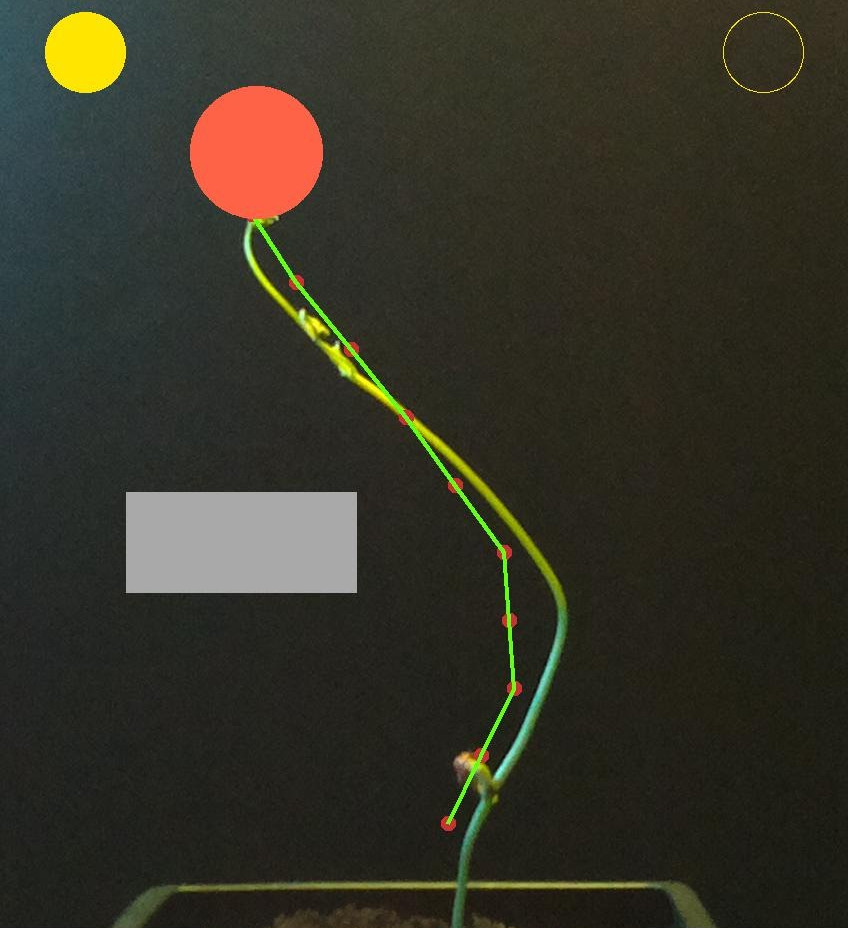}
   }
   \caption{\label{fig:results:reality} Sequence of images showing the course of a reality-gap experiment. \normalfont The yellow circles on top indicate which light is on (filled: on, empty: off).
The larger filled red circle is the target area and the gray rectangle is the obstacle the plant is not allowed to touch.
}
\end{figure*}

To test controller performance in the physical world, we examine whether it can guide a natural bean plant around a virtual obstacle without colliding, and reach the target area.
This addresses the reality-gap problem~\citep{koos13}, which states that controllers evolved in simulation do not always transfer to a real setup, because the simulation is limited in principle.
To test the reality gap, we use the setup described in Sec.~\ref{sec:biosetup}.
Computer vision detects the stem, feeding into the ANN evolved in simulation, which controls light stimuli provided to the real bean plant.
The controller we select is evolved in the \textit{left target} experiment, with the obstacle centered ${\approx}6.3$~cm left of the plant anchor, $12.5$~cm above the soil.
The bio-hybrid setup completed the task with the real plant, although the plant's circumnutation behavior brought it in close proximity to the obstacle (see Fig.~\ref{fig:results:reality} and video\footnote{Find a video at: \url{https://vimeo.com/265144652}}.
In the experiment, the controller initially maintained the right light, guiding the plant away from the obstacle for ${\approx}37$~h, until the plant was $7.4$~cm right of the plant origin and $15.4$~cm above the soil, see Fig.~\ref{fig:results:reality:2}.
Then it switched to the left light, quickly bringing the plant tip to the opposite side, while the stem retained some stiffness.
After less than $2$~h, the plant tip is roughly in the center, with a pronounced curve in the stem, see Fig.~\ref{fig:results:reality:3}, as lower tissues already stiffen.
Then follows a phase of quick light alterations, as the controller guides the plant tip close to the obstacle edge, leaving it near to the target after clearing the obstacle. The left light is then triggered for another $37$~h, successfully guiding the plant to the target, while the curvature of the stiffened stem allows it to entirely avoid the obstacle. The controller's effectiveness in exploiting the plant's stiffening behavior is seen by comparing this result to those of~\citep{Hofstadler2017taas}. Here, stiffening has resulted in noticeable stem curvature, while in~\citep{Hofstadler2017taas} the stems have a straight shape, even after being steered to targets on opposing sides.
The evolved controller together with the real plant achieves $92.4$\%~fitness,
see Fig.~\ref{fig:results:reality:4}.
The experiment was repeated two further times, achieving fitnesses of $92.0$\% and $87.7$\%.
In the latter, the bean grows abnormally.
It is significantly slower (by half) than the others, which have comparable growth speeds to those in~\citep{wahby15, Hofstadler2017taas,wahby2016evolutionary}.
While both other experiments last $75$~h, this bean only grows above the obstacle in hour~$84$.
However, the controller behaves correctly, and the bean approaches the target until hour~$198$, when its preexisting anomalies cause collapse.
We record the fitness when the plant tip reaches the target area, because it is difficult to hit a single point, considering the stiffened tissues. However, it is possible to achieve higher fitness values, as a maximum fitness of 99.3\% was later observed in the second experiment.

\section{Discussion and Conclusion}

Following the objective of using natural plants to do additive manufacturing and to implement a real-world, tunable developmental system, we have set up a toolchain to shape natural plants in an evolutionary robotics approach. We acquired data about the growth and motion of a plant, trained a state-of-the-art LSTM as plant model, evolved robot controllers using the LSTM as simulator, and successfully tested these controllers for the reality gap. Our focus was on the delicate interplay of plant motion and tissue-stiffening to shape plants around obstacles with collision-free control. Early on, the plant motion has to be controlled strategically to provoke the correct stiffened shape later. We call this particular phenomenon `embodied memory' because the plant tip motion and the orientation of the whole plant during the experiment is integrated over time and partially reflected in the final stiffened shape of the plant. This is particularly different from other tasks in evolutionary robotics, such as the navigation of a mobile robot, where the full history (robot trajectory) has only a minor and indirect influence on the task completion. The task could arguably be compared to the control of a robot arm where joints closer to the base lose their flexibility over time.
In comparison to 
our previous work~\citep{Hofstadler2017taas}, where controllers were evolved to guide the plant tip into randomly generated targets, our target control behavior is evidently more complex, because the evolutionary process required 300 more generations till convergence.
The controller here needs to be aware of the whole plant body instead of only the tip, in order to be able to avoid hitting the obstacle at any point along the stem. 

A key achievement of this work is the successful application of methods from machine learning (LSTM network) to create a holistic plant model. Unfortunately, such models representing the plant's macroscopic reactions to stimuli are not readily available from plant science. We have shown that with data from a few generic plant experiments, a sufficient model can be obtained. However, the limited availability of data is a challenge as common in machine learning and especially deep learning. Growing plants as such can be parallelized but considerable costs are added by controlling the light conditions and tracking, hence data is sparse. We have reported our approach to data augmentation, which may also have some potential to scale up.

The presented methodology with heavy use of machine learning techniques has potential to scale up to more desirable growth tasks that go beyond mere obstacle avoiding. Options are to grow plant patterns on meter-scales or more, to grow and control multiple plants within the same area, and to grow also 3D-patterns. Besides their natural aesthetics these grown shapes may also have functionality, for example, as architectural artifacts (green walls, roofs, etc.).
Therefore, we plan to make the transition to a 3D setup in future work, where we can grow more complex shapes, such as spirals, geometrical objects, or even writing. Controlling multiple plants concurrently will also add complexity, especially once we allow them to interact. We plan to automatically braid plants, use them to change material properties in construction, to investigate different plant species, and to grow complex structures, such as meshes or even benches. In addition, we investigate options to use phytosensing (i.e., using plants as sensors) that could help to implement synergistic robot-plant interactions.
Hence, the presented machine learning approach of shaping plants opens doors for autonomous bio-hybrid systems with promising applications.

\section*{Acknowledgements}
Project `{\it flora robotica}' has received funding from the European Union's
Horizon 2020 research and innovation program under the FET grant
agreement, no.~640959.

\bibliographystyle{plainnat}
\bibliography{bib}

\begin{thebibliography}{27}
\providecommand{\natexlab}[1]{#1}
\providecommand{\url}[1]{\texttt{#1}}
\expandafter\ifx\csname urlstyle\endcsname\relax
  \providecommand{\doi}[1]{doi: #1}\else
  \providecommand{\doi}{doi: \begingroup \urlstyle{rm}\Url}\fi

\bibitem[Abadi et~al.(2016)]{abadi2016tensorflow}
M.~Abadi et~al.
\newblock Tensorflow: A system for large-scale machine learning.
\newblock In \emph{OSDI}, volume~16, pages 265--283, 2016.

\bibitem[Bastien et~al.(2015)]{bastien2015unified}
R.~Bastien et~al.
\newblock A unified model of shoot tropism in plants: Photo-, gravi- and
  propio-ception.
\newblock \emph{PLoS Comput Biol}, 11\penalty0 (2), 2015.

\bibitem[Bongard(2013)]{bongard13}
J.~C. Bongard.
\newblock Evolutionary robotics.
\newblock \emph{Communications of the ACM}, 56\penalty0 (8):\penalty0 74--83,
  2013.

\bibitem[Checa et~al.(2008)]{checa2008mapping}
O.~E. Checa et~al.
\newblock Mapping {QTL} for climbing ability and component traits in common
  bean ({\it phaseolus vulgaris l.}).
\newblock \emph{Molecular Breeding}, 22\penalty0 (2), 2008.

\bibitem[Chervenski et~al.(2012)]{multineat}
P.~Chervenski et~al.
\newblock {MultiNEAT}, 2012.
\newblock URL \url{http://www.multineat.com/}.

\bibitem[Chollet et~al.(2015)]{chollet2015keras}
F.~Chollet et~al.
\newblock Keras.
\newblock 2015.
\newblock URL \url{https://keras.io/}.

\bibitem[Christie et~al.(2013)]{christie2013phototropism}
J.~M. Christie et~al.
\newblock Shoot phototropism in higher plants: New light through old concepts.
\newblock \emph{American Journal of Botany}, 100\penalty0 (1):\penalty0 35--46,
  2013.

\bibitem[Garz{\'o}n and Keijzer(2011)]{garzon2011plants}
P.~C. Garz{\'o}n and F.~Keijzer.
\newblock Plants: Adaptive behavior, root-brains, and minimal cognition.
\newblock \emph{Adaptive Behavior}, 19\penalty0 (3):\penalty0 155--171, 2011.

\bibitem[Graves et~al.(2013)]{graves2013speech}
A.~Graves et~al.
\newblock Speech recognition with deep recurrent neural networks.
\newblock In \emph{IEEE Int. Conf. on Acoustics, speech and signal processing
  (ICASSP)}. IEEE, 2013.

\bibitem[Halloy et~al.(2007)]{HalloyEtAl07}
J.~Halloy et~al.
\newblock Social integration of robots into groups of cockroaches to control
  self-organized choices.
\newblock \emph{Science}, 318\penalty0 (5853):\penalty0 1155--1158, November
  2007.

\bibitem[Hamann et~al.(2015)]{hamann2015florarobotica}
H.~Hamann et~al.
\newblock {\it flora robotica} -- mixed societies of symbiotic robot-plant
  bio-hybrids.
\newblock In \emph{Proc. of IEEE Symposium on Computational Intelligence}.
  IEEE, 2015.

\bibitem[Hamann et~al.(2017)]{hamann2017flora}
H.~Hamann et~al.
\newblock Flora robotica--an architectural system combining living natural
  plants and distributed robots.
\newblock \emph{arXiv:1709.04291}, 2017.

\bibitem[Hochreiter and Schmidhuber(1997)]{hochreiter1997long}
S.~Hochreiter and J.~Schmidhuber.
\newblock Long short-term memory.
\newblock \emph{Neural computation}, 9\penalty0 (8):\penalty0 1735--1780, 1997.

\bibitem[Hofstadler et~al.(2017)]{Hofstadler2017taas}
D.~N. Hofstadler et~al.
\newblock Evolved control of natural plants: Crossing the reality gap for
  user-defined steering of growth and motion.
\newblock \emph{ACM Trans. Auton. Adapt. Syst. (TAAS)}, 12\penalty0
  (3):\penalty0 15:1--15:24, 2017.

\bibitem[Koos et~al.(2013)]{koos13}
S.~Koos et~al.
\newblock The transferability approach: Crossing the reality gap in
  evolutionary robotics.
\newblock \emph{IEEE Trans. on Evo. Comp.}, 17\penalty0 (1):\penalty0 122--145,
  2013.

\bibitem[Lindenmayer(1975)]{lindenmayer75}
A.~Lindenmayer.
\newblock Developmental algorithms for multicellular organisms: A survey of
  {L}-systems.
\newblock \emph{Journal of Theoretical Biology}, 54\penalty0 (1):\penalty0
  3--22, 1975.

\bibitem[Mansouryar et~al.(2012)]{mansouryar2012smoothing}
M.~Mansouryar et~al.
\newblock Smoothing via iterative averaging (sia) a basic technique for line
  smoothing.
\newblock \emph{Int. Jo. of Computer and Electrical Eng.}, 4\penalty0 (3),
  2012.

\bibitem[Namin et~al.(2017)]{namin2017deep}
S.~T. Namin et~al.
\newblock Deep phenotyping: Deep learning for temporal phenotype/genotype
  classification.
\newblock \emph{bioRxiv}, 2017.

\bibitem[Nelson et~al.(2009)]{nelson_2009_fitness}
A.~L. Nelson et~al.
\newblock Fitness functions in evolutionary robotics: {A} survey and analysis.
\newblock \emph{Robotics and Autonomous Systems}, 57:\penalty0 345--370, 2009.

\bibitem[Stanley et~al.(2004)]{stanley04}
K.~O. Stanley et~al.
\newblock Competitive coevolution through evolutionary complexification.
\newblock \emph{Journal of Artificial Intelligence Research}, 21\penalty0
  (1):\penalty0 63--100, 2004.

\bibitem[Sutskever et~al.(2014)]{sutskever2014sequence}
I.~Sutskever et~al.
\newblock Sequence to sequence learning with neural networks.
\newblock In \emph{Advances in neural information processing systems}, pages
  3104--3112, 2014.

\bibitem[Wahby et~al.(2015)]{wahby15}
M.~Wahby et~al.
\newblock Evolution of controllers for robot-plant bio-hybdrids: A simple case
  study using a model of plant growth and motion.
\newblock In \emph{Proc. of the 25th Workshop on Computational Intelligence},
  pages 67--86. KIT Scientific Publishing, 2015.

\bibitem[Wahby et~al.(2016)]{wahby2016evolutionary}
M.~Wahby et~al.
\newblock An evolutionary robotics approach to the control of plant growth and
  motion.
\newblock In \emph{IEEE 10th Int. Conf. on Self-Adaptive and Self-Organizing
  Systems (SASO)}, pages 21--30. IEEE, 2016.

\bibitem[Watson et~al.(2002)]{watson02}
R.~A. Watson et~al.
\newblock Embodied evolution: Distributing an evolutionary algorithm in a
  population of robots.
\newblock \emph{Robotics and Autonomous Sys.}, 39\penalty0 (1), 2002.

\bibitem[Zahadat et~al.(2014)]{zahadat14}
P.~Zahadat et~al.
\newblock Social adaptation of robots for modulating self-organization in
  animal societies.
\newblock In \emph{IEEE Int. Conf. on Self-Adaptive and Self-Organizing Systems
  Workshops (SASOW)}, pages 55--60, Sept 2014.

\bibitem[Zahadat et~al.(2017)]{zahadat2017a}
P.~Zahadat et~al.
\newblock Vascular morphogenesis controller: A generative model for developing
  morphology of artificial structures.
\newblock In \emph{Proceedings of the Genetic and Evolutionary Computation
  Conference}, GECCO '17, pages 163--170, 2017.

\bibitem[Zamuda et~al.(2014)]{zamuda14}
A.~Zamuda et~al.
\newblock Vectorized procedural models for animated trees reconstruction using
  differential evolution.
\newblock \emph{Information Sciences}, 278:\penalty0 1--21, 2014.

\end{thebibliography}

\end{document}